\documentclass[fleqn,10pt]{wlscirep} \usepackage{float}

\title{Urban morphology meets deep learning: Exploring urban forms in one million cities, town and villages across the planet}

\author{Vahid Moosavi} \affil{Chair for Computer Aided Architectural Design, Department of Architecture, ETH Zurich, 8093, Switzerland} \affil{svm@arch.ethz.ch}

\begin{abstract}
	Study of urban form is an important area of research in urban planning/design that contributes to our understanding of how cities function and evolve. However, classical approaches are based on very limited observations and inconsistent methods. As an alternative, availability of massive urban data collections such as Open Street Map from the one hand and the recent advancements in machine learning methods such as deep learning techniques on the other have opened up new possibilities to automatically investigate urban forms at the global scale. 
	
	In this work for the first time, by collecting a large data set of street networks in more than one million cities, towns and villages all over the world, we trained a deep convolutional auto-encoder, that automatically learns the hierarchical structures of urban forms and represents them via dense and comparable vectors. We showed how the learned urban vectors could be used for different investigations. Using the learned urban vectors, one is able to easily find and compare similar urban forms all over the world, considering their overall spatial structure and other factors such as orientation, graphical structure, and density and partial deformations. Further cluster analysis reveals the distribution of the main patterns of urban forms all over the planet. 
\end{abstract}
\begin{document}

\flushbottom \maketitle

\thispagestyle{empty}

\section*{Introduction} 
\begin{figure}
	[!htb] \centering \includegraphics[width=300pt, height=100pt]{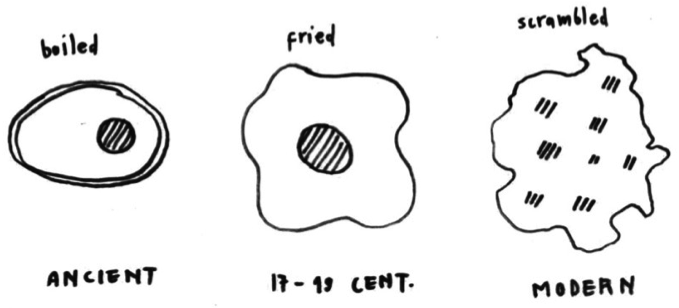} \caption{Qualitative and conceptual urban models: Analogy of urban forms with an egg by Cedric Price} 
\label{fig:Fig1} \end{figure}

Cities are among the most complex cultural products that are expanded all over the planet. Although each city has its own unique story, study of cities at the global scale has been an important area of research for a long time. Within the domain of urban studies, urban morphology is the classical study of urban forms and their underlying formation processes and forces over time. Historically, urban morphologist study cities based on qualitative approaches and personal observations and limited to few famous cities or even one specific location\cite{salat2011cities,lynch1984good,lynch1960image}. This approach usually leads to qualitative and conceptual urban models such as those tripartite models of Cedric Price’s analogy of cities with three different ways of cooking eggs (see Figure \ref{fig:Fig1} ), identified by medieval cities with concentric patterns, industrial cities with mechanical grids and ecological cities with polycentric patterns and organic forms\cite{shane2005recombinant}. 

As another example of qualitative studies of urban forms and street networks at the neighborhood level one can refer to famous archetypical models such gridirons, fragmented parallel, warped parallel, loops and lollipops and lollipops on a stick\cite{southworth2013streets}. Next to qualitative studies, recently there have been several attempts to study urban forms based on quantitative methods, usually based on complexity theory or network science point of view or the so called city science among the community of physicists\cite{barthelemy2008modeling,buhl2006topological,cardillo2006structural,masucci2009random,strano2013urban,fiasconaro2016spatio}. Most of these studies are either based on theoretical models of urban morphology such as fractals\cite{batty2008size} or based on very limited samples of city networks, usually less than 100 networks\cite{barthelemy2008modeling,buhl2006topological,cardillo2006structural,strano2013urban,louf2014typology}. Further, most of these quantitative studies, are trying to generalize their findings into universal equations in terms of scaling laws or distributions of size and densities of the built environments. A complete list of these quantitative measures of urban forms can be found at\cite{boeing2017measuring}. 

To sum up, as Boeing\cite{boeing2017osmnx} argues, current quantitative studies suffer from four main limitations of small sample sizes, excessive network simplification, difficult reproducibility, and the lack of consistent and easy-to-use research tools. In addition to these limits, comparing to qualitative approaches, one of the main critiques to current quantitative approaches in urban morphology is that they sometimes loose the big picture by the use of detailed analytical metrics, while the overall visual and qualitative patterns are also very important for architects and urban designers. 

\begin{figure}
	[!htb] \centering \includegraphics[width=\linewidth,height=200pt]{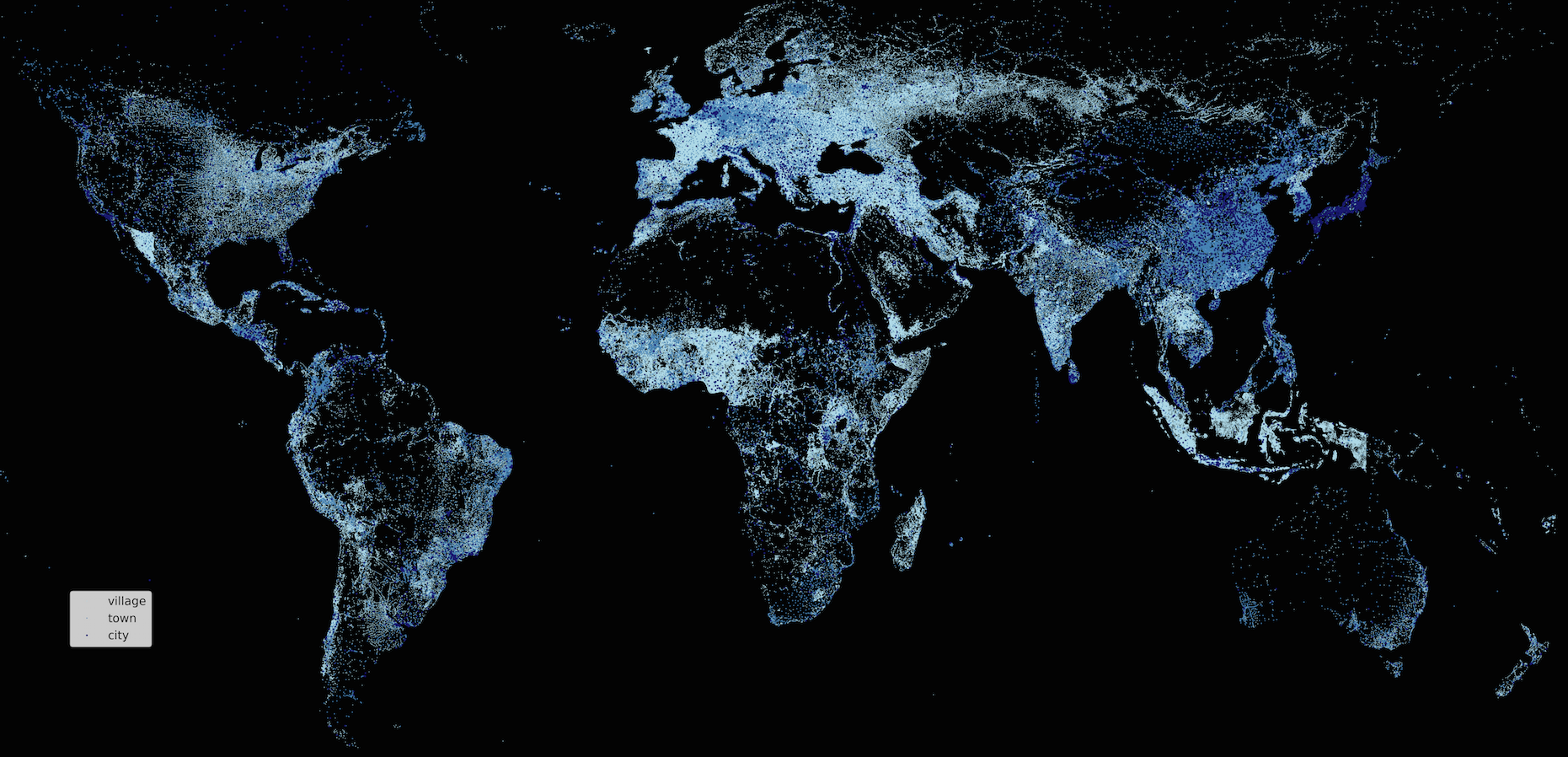} \caption{Data from more than one million cities, towns and villages that are indexed in Open Street Map is available as a free Big Data set on cities and built environments all over the world.} 
\label{fig:Fig2} \end{figure}

At the same time, nowadays cities are producing enormous amounts of data about different aspects of life in our planet. If we couple these different sources together, they will reveal very interesting aspects of cities and the ways they function. One of the great sources of urban data is Open Street Map (OSM), which is an open source project that provides a free and publicly available map of the whole planet. Figure \ref{fig:Fig2} shows the spatial distribution of more than one million indexed locations across the planet. 

In this work, our goal is to investigate the potential benefits of this large collection of urban data for architects and urban design researchers. To the best of our knowledge this is the first time that we investigate the urban forms of more than one million locations across the planet. Ultimately we would like to develop a search engine of urban forms that automatically indexes millions of locations and makes it possible for the researchers and practitioners to explore and investigate different aspects of urban forms all over the planet. From the point of view of machine learning this can be framed as an unsupervised learning task such as feature learning and data-clustering, where for example, one is interested to identify the main urban patterns based on their spatial form or supervised learning tasks, where one is interested to study urban forms regarding to other urban qualities such as traffic, population health or urban pollution. In this paper, we only report the results of our first step toward clustering and pattern recognition of urban forms.

Technically speaking, similar to texts or images, urban forms have complex and hierarchical structures, composed of building layouts, street segments, neighborhood layouts and main street networks. Therefore, while network based analytics is the common approach in quantitative urban studies, as we will show later considering the inherent complexity of urban patterns, the use of the so-called representation learning and deep learning algorithms\cite{bengio2013representation} such as Convolutional Neural Networks (CNNs)\cite{fukushima1982neocognitron,lecun1990handwritten} in computer vision and natural language processing seems very promising. Similar to the idea of Word2vec project\cite{mikolov2013distributed}, where each word gets a unique representation in vector space, after training a deep learning model, each unique urban form gets a vector based representation, where similar urban forms gets similar representations based on their similarities in density, orientation and overall structures, regardless of translations, partial deformations and rotations. Next, we use the learned vectors for further analyses such as clustering and pattern recognition.

In the next section, we first explain our experimental set up and the data collection pipeline from OSM. Next, we describe the architecture of the deep learning model we used in this work. Then, we describe the main results and finally we discuss possible next steps and conclude the paper.

\section*{Methods}

\subsection*{Experimental set up and data collection} While, OSM is a wiki-style project, based on several studies\cite{corcoran2013analysing,arsanjani2015introduction,barron2014comprehensive,girres2010quality,haklay2010good} it has a relatively high quality data set in terms of coverage of the whole planet. At the time of our data collection, according to the tags of indexed places (See \href{https://taginfo.openstreetmap.org/keys/place#values}{taginfo}.) there were 8,325 cities, 84,785 towns and 1,036,494 villages, totally 1,130,591 locations indexed in OSM. OSM freely provides data sets as a special form of XML via its APIs such as Overpass or in bulk mode from several different sources such as GEOFABRIC. These data sets can be converted to shape files or standard graphical formats\cite{boeing2017osmnx}. 

While ultimately it would be interesting to compare the urban forms via graph-based kernels and CNN models with non-Euclidean kernels\cite{defferrard2016convolutional}, since graphical methods focus on connectivity and the topology of the networks, they might loose the information on spatial and visual patterns. Further, converting OSM data of more than a million locations to graphs and the graph analysis processes can be time consuming or depending on the computational power almost infeasible in a reasonable time. One quick solution is to take the images of the cities at the same spatial scale and use them to train a standard image based CNN. Another advantage of using image representation over the graphical representations of urban forms is that it is easily possible to combine different aspects of cities such as buildings, streets and land use information next to each other in one single image, while they can not be easily incorporated into one single graphical representation.

\begin{figure}
	[!htb] \centering \includegraphics[width=300pt,height=300pt]{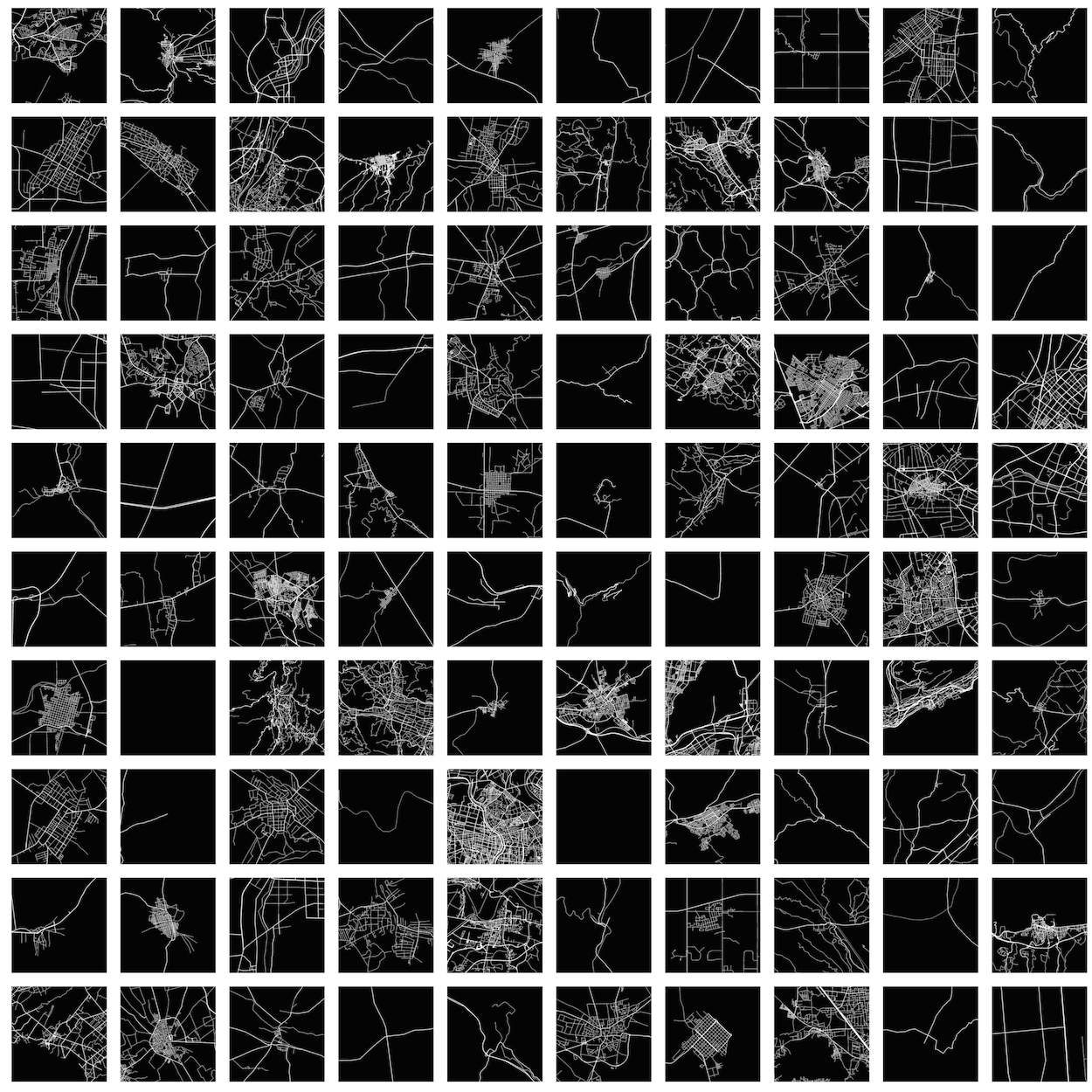} \caption{Images of urban forms at randomly selected cities, towns or villages, all with the same spatial scale} 
\label{fig:Fig3} \end{figure}

Next, while the immediate idea would be to use the satellite images, since we were interested only in urban forms, we designed our images with specific color codes and styles across the planet, using Mapbox studio web application. Mapbox studio is a web application for map styling, where one can easily design a visual map and select few specific layers to be shown on the map with specific styles including the thicknesses of the lines and the colors of polygons. Therefore, a nice and easy solution for us would be to create a styled map, where for example the roads are red, buildings are blue and the greeneries are green. Further, the thickness of the lines specifies the primary, secondary and tertiary roads as well as their types such as highways, railroads and tunnels, etc. Next, by using the static API of Mapbox and providing the geographic coordinates of the locations that were extracted from OSM we downloaded the images of each urban form, with the same spatial scale (i.e. zoom level). However, unfortunately, the building information (i.e. the building footprints) is missing in many locations all over the world. Therefore, in this work, we only focused on road networks. Note that also it is possible that for some locations there is no spatial information and as a result the static API returns an empty file, which will be removed from the training data sets. Since we fixed the spatial scale of images, it is logically expected that while a single image can easily capture one village, with the same size, we only can capture a small part of a large city such as Beijing or London. Nevertheless, in this experiment, we only took one image from the center of each location, regardless of its overall size. Finally, we collected 962,639 images. The initial images have the size of 2400x2400x1, which are very large for typical deep learning applications. Therefore, after cropping, we resized them to 256x256x1 binary images. Figure \ref{fig:Fig3} shows few examples of the styled maps in several randomly chosen locations. 
In the next section we describe the architecture of the deep learning model we used in this study.

\subsection*{Deep learning model} Recently, CNNs\cite{fukushima1982neocognitron,lecun1990handwritten} have changed the landscape of computer vision applications\cite{krizhevsky2012imagenet,he2016deep} as they efficiently learn complex hierarchical patterns in image data sets or in general in any homogenous type of data such as texts or different time series related applications. Although, it is usually common to train CNNs in supervised learning tasks such as classification with labeled data sets, here we do not have any labeled data set. In this case the usual scenario is to use Auto Encoder (AE) networks\cite{hinton2006reducing,huang2007unsupervised}, where the model learns to minimize its reconstruction loss. In other words, the network learns to predict its input. An AE has two main parts of encoder and decoder, where the input of the encoder will be the target output of the decoder. Usually the activations of the trained middle layer (i.e. the last layer of the encoder) can be used as a dense vector based representation of the inputs. Then, these vectors can be used for further tasks such as learning a low dimensional embedding or data clustering. While, the original AE networks are based on stacks of fully connected layers, they do not easily scale with high dimensional images. Therefore, a common choice of architecture is to combine convolutional kernels in AE architecture, known as Convolutional Auto Encoders (CAE)\cite{huang2007unsupervised,masci2011stacked}. CAEs scale very well to high dimensional images comparing to fully connected AEs, which require a relatively larger number of parameters to be learned. Further, CAEs consider the 2D structure of images, while in fully connected AEs one needs to flatten the images to a high dimensional vector, where each pixel is one dimension. This introduces a redundancy in the parameters of the model, as all the features need to play globally in all the dimensions. This is opposite to CAEs, where different kernels learn to focus on certain patterns.

In our work, we trained a CAE with tied weights with convolution layers in the encoder layers and the de-convolution (or transposed convolutions)\cite{zeiler2010deconvolutional} in the decoder layers with no fully connected layers. It is very common to alternate convolution layers with max-pooling layers in the context of classifications. However, replacing max-pooling layers with strides will also give competitive results\cite{springenberg2014striving}. In our work, we chose the stride of 2 in all the layers, which divides the widths and lengths of images to two after each layer in the encoder part and vice versa in the decoder part. Also, we added ReLu nonlinearity on top of the output of each convolution operation. We tested several architectures and finally we chose a network with 5 encoder layers with 15,15,15,10,10 kernels in each layer correspondingly. For all the layers we used the kernels with the size of 5x5. We used an implementation of this algorithm in TensorFlow environment with CPU. A complete training cycle with batch size of 50 and 50 epochs took around a week on an IMAC 5k.

\section*{Results}

\subsection*{Reconstruction quality of the trained auto-encoder} 
\begin{figure}
	[!htb] \centering \includegraphics[width=300pt,height=100pt]{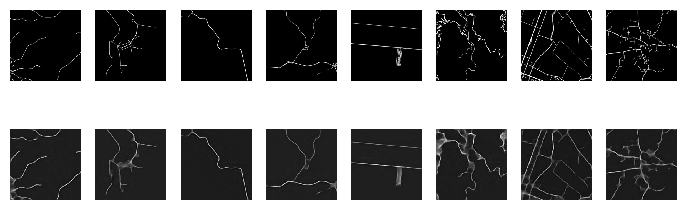} \caption{Original input (top) and the reconstructed (bottom) images by the trained convolutional auto-encoder} 
\label{fig:Fig4} \end{figure}

After training the network, we also visually inspected the reconstruction quality of the trained auto-encoder for several random images. Figure \ref{fig:Fig4} shows several reconstructed images in comparison to their original images. As we can see, while the model is able to reconstruct the overall image, it misses some fine-grained details at the neighborhood levels. 
Next, we projected all the images to the trained network and saved the activations of the last encoder layer that gave us an 8x8x10 matrix for each input image. After unrolling each matrix we had a 640 dimensional vector for each image. This gives us a dense representation of the high dimensional input images (from originally 256x256x1 or 65,536 dimensions to 8x8x10 or 640 dimensions). Hereafter, we call these dense vectors urban vectors. In principle, while the notion of similarity between two urban forms is hard to quantify, we expect that the learned urban vectors represent the hierarchical structure of the underlying road networks and capture the overall urban form considering the important issues such as density, deformations, orientations, translation and partial rotations. 
\subsubsection*{Urban form exploration} 
\begin{figure}
	[!htb] \centering \includegraphics[width=300pt, height=250pt ]{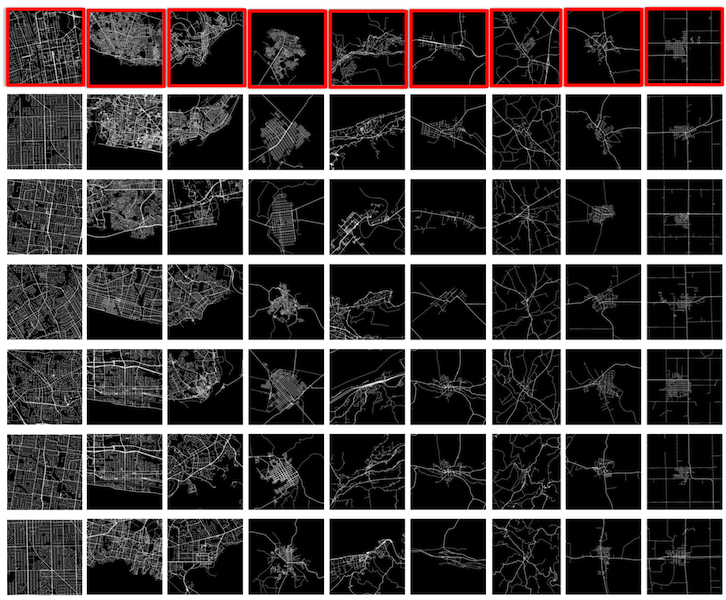} \caption{Finding the 6 most similar urban forms using the learned urban vectors } 
\label{fig:Fig5} \end{figure}

Similar to Word2Vec\cite{mikolov2013distributed}, where similar words have similar dense vector based representations, here in the first step we used the learned urban vectors to find similar urban forms. Fitting a KNN model to the urban vectors, we achieved very satisfying results. In Figure \ref{fig:Fig5} each column shows the 6 most similar urban forms to the chosen urban form in the first cell of that column. Here, the similarity is defined based on the Euclidean distance between the urban vectors. As expected, the learned urban vectors nicely represent the visual urban forms, considering their density, main structure, translation, orientation, and partial deformations. However, with the use of normal convolutional kernels, the model does not fully learn the rotation invariances. 
\subsubsection*{A spectrum of urban forms across the planet} 
\begin{figure}
	[!htb] \centering \includegraphics[width=\linewidth, height=200pt ]{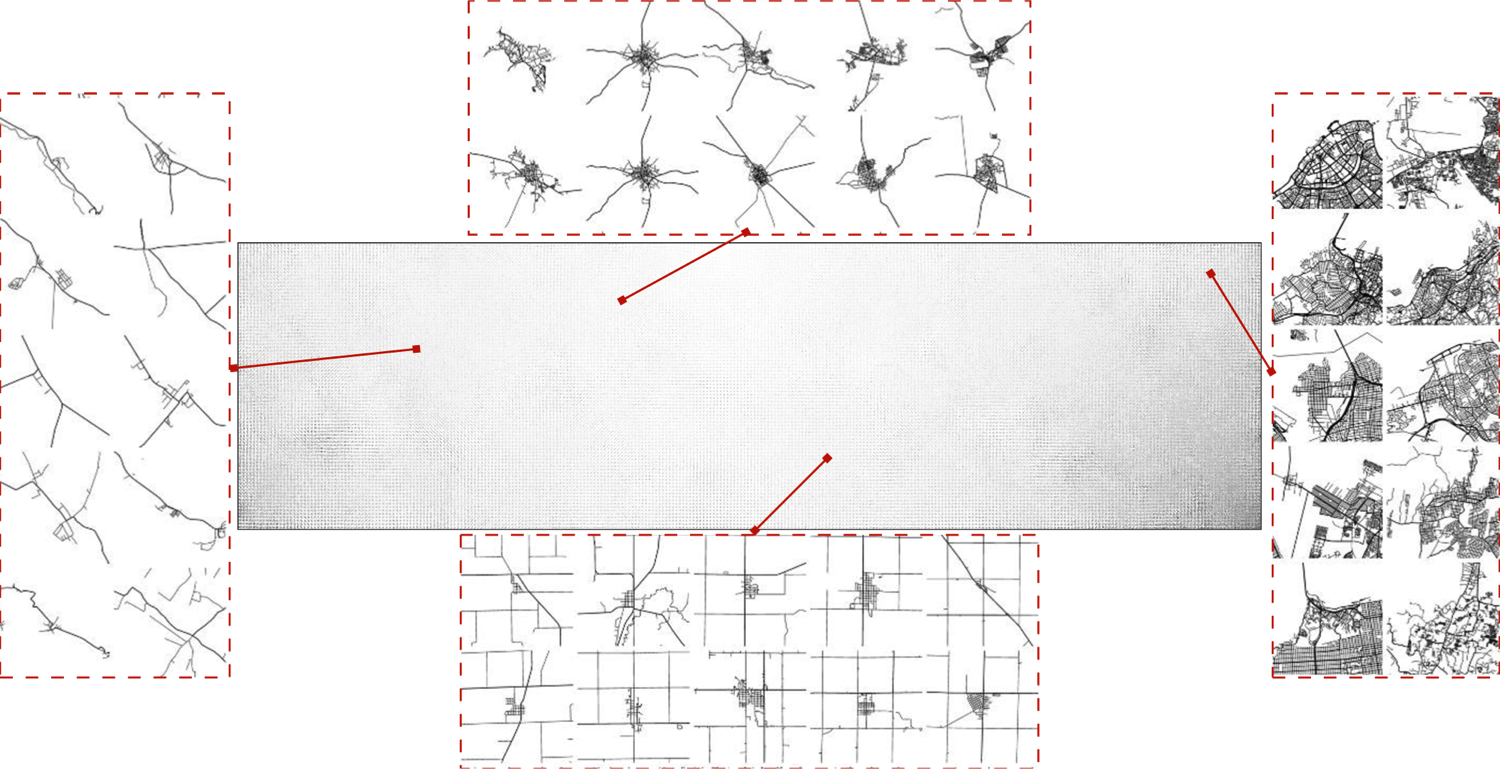} \caption{An automatically generated spectrum of urban development patterns for around 1 million cities, towns and villages across the planet} 
\label{fig:Fig6} \end{figure}

Next, using the urban vectors of all the cities, we trained a large Self Organizing Map (SOM)\cite{kohonen1998self}, which projects the 640 dimensional urban vectors to a regular 2D grid. The common option for the dimensionality reduction and visualization is T-SNE\cite{maaten2008visualizing}. However, T-SNE is very slow with large and high dimensional data sets. Opposite to T-SNE, since SOM is performing the data reduction and dimensionality reduction at the same time, it scales very well with the size of the data. Figure \ref{fig:Fig6} shows a section of an automatically generated spectrum of urban forms all over the world. As four zoomed areas of this map show similar urban forms are placed next to each other in a way that overall they create a very smooth spectrum of urban forms all over the planet. An interactive version of this spectrum can be found at \href{https://sevamoo.github.io/cityastext/}{the project website}. 

\subsection*{Analyzing the clusters of urban forms} Next to visual explorations of the urban forms, we clustered all the urban forms via the learned urban vectors. Toward this goal we used SOM algorithm, which has been shown to be a powerful clustering algorithm. One unique benefit of SOM for data clustering is that unlike other clustering algorithms, it labels the clusters in a way that similar clusters get similar indices. We trained a SOM with a one-dimensional grid topology that indexes the high dimensional vectors into an ordered series of one-dimensional indexes\cite{moosavi2014computing,moosavi2017contextual}. Each index (i.e. each node of the trained SOM) will represent a cluster of urban forms, while similar indices are referring to similar high dimensional urban forms. Therefore, in addition to clustering and data reduction the trained SOM creates an intuitive spectrum of urban forms, which can be easily visualized in a colored geo-map\cite{moosavi2017contextual}. (See Figure \ref{fig:Fig8} and Figure \ref{fig:Fig10}.) Further, since the labels are comparable, we can study the distribution of main clusters of urban forms in an intuitive manner. 

\begin{figure}
	[!htb] \centering \includegraphics[width=200pt, height=150pt ]{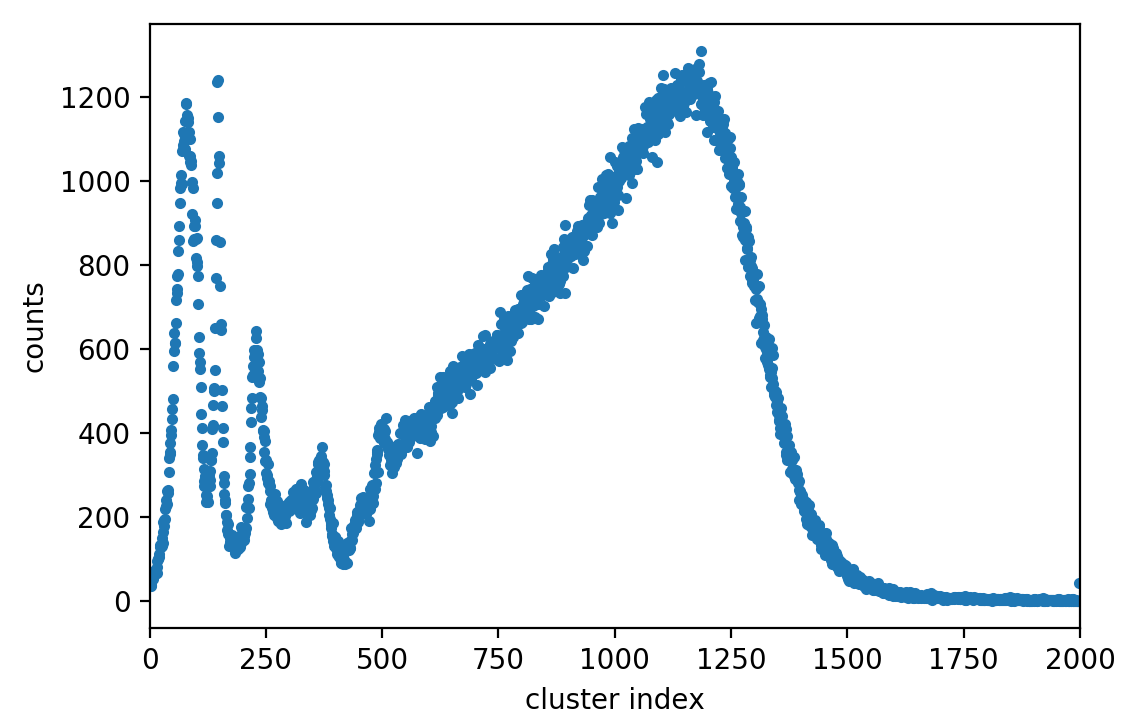} \caption{Distribution of urban forms within 2000 identified clusters} 
\label{fig:Fig7} \end{figure}

\begin{figure}
	[!htb] \centering \includegraphics[width=\linewidth,height=200pt]{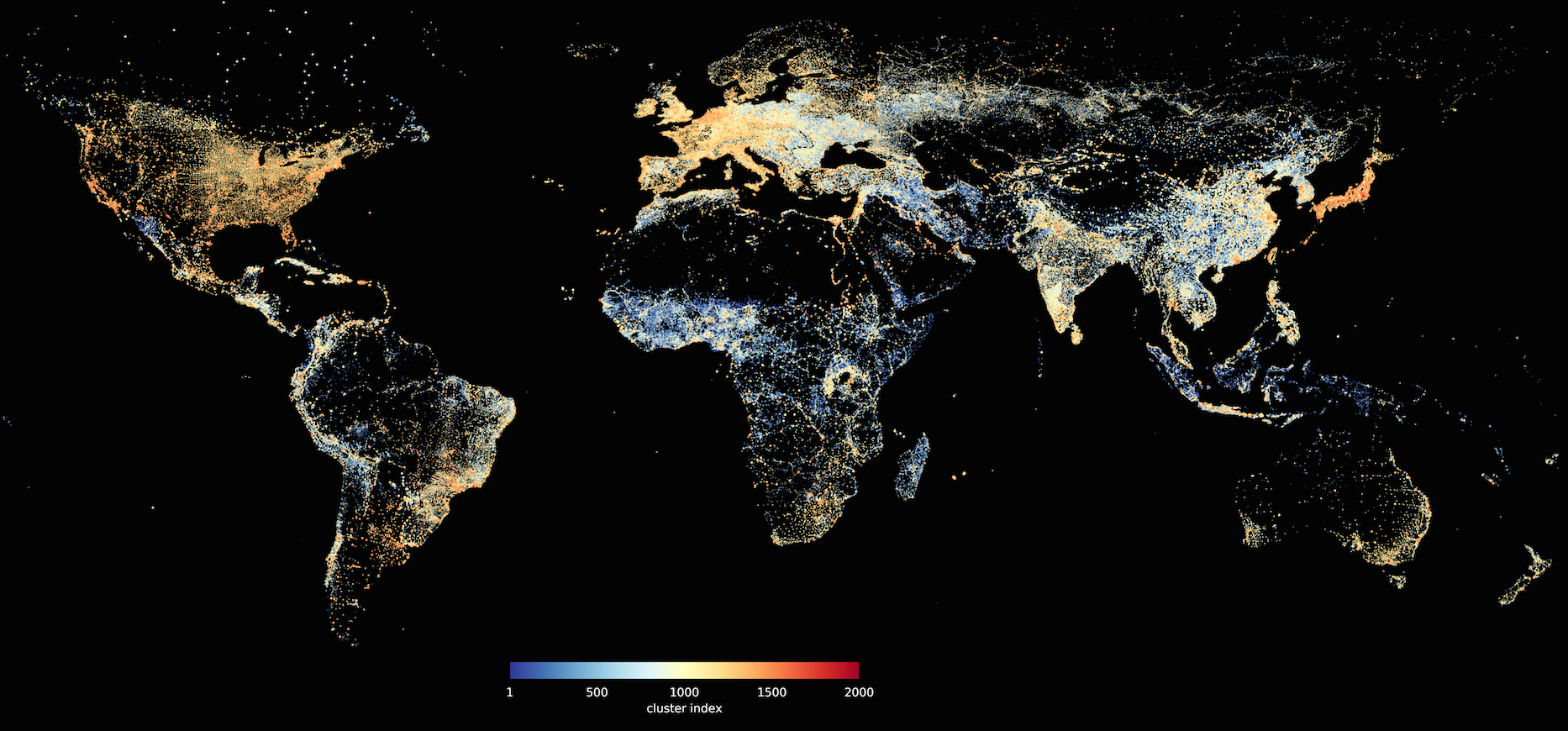} \caption{Global distribution of urban forms: Similar colors refer to similar urban patterns} 
\label{fig:Fig8} \end{figure}

\begin{figure}
	[!htb] \centering \includegraphics[width=200pt, height=200pt]{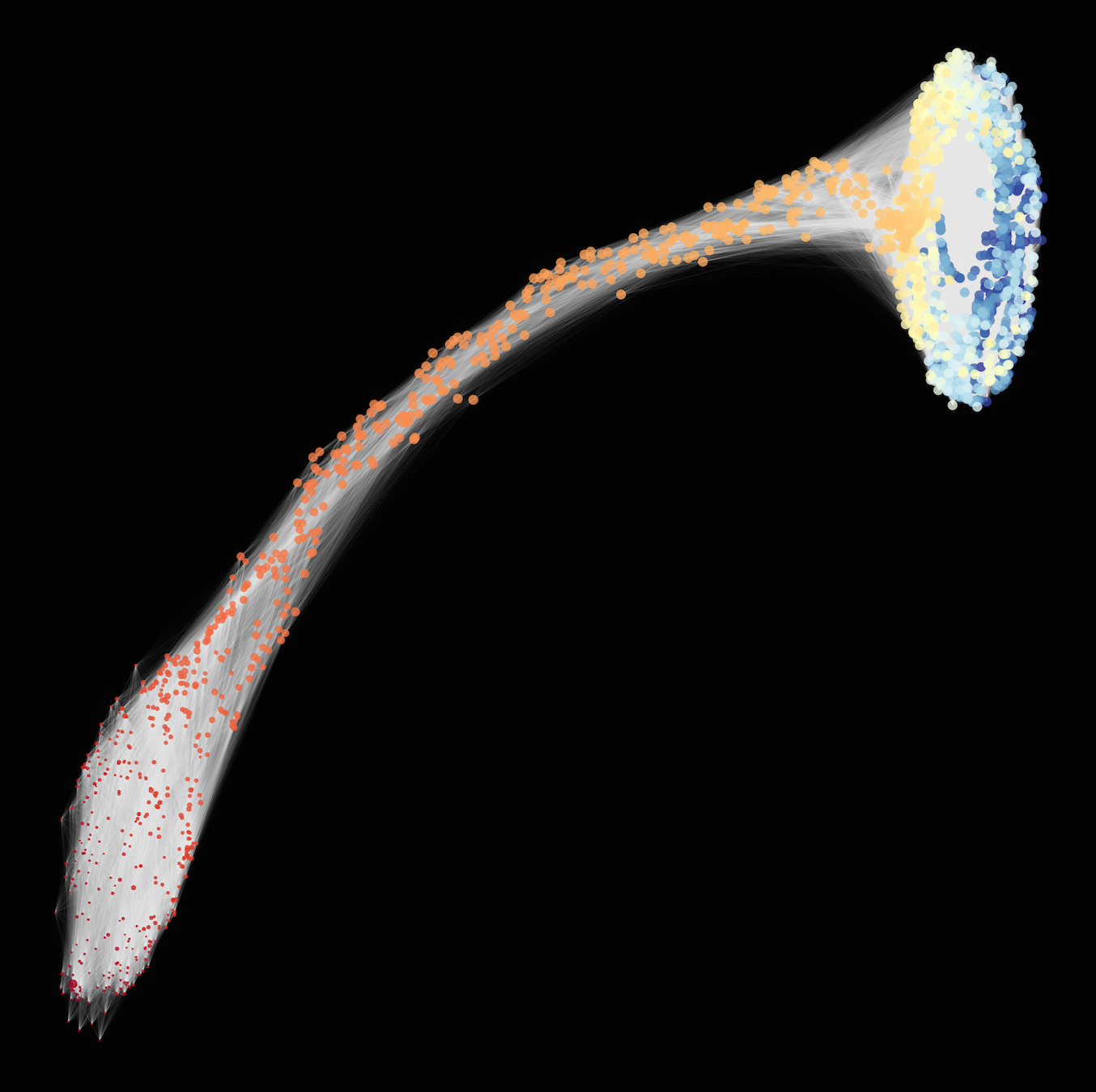} \caption{The topology of urban forms: each node represents a cluster of urban forms. similar gets are closer to each other and have similar color codes.} 
\label{fig:Fig9} \end{figure}

\begin{figure}
	[!htb] \centering \includegraphics[width=300pt, height=200pt]{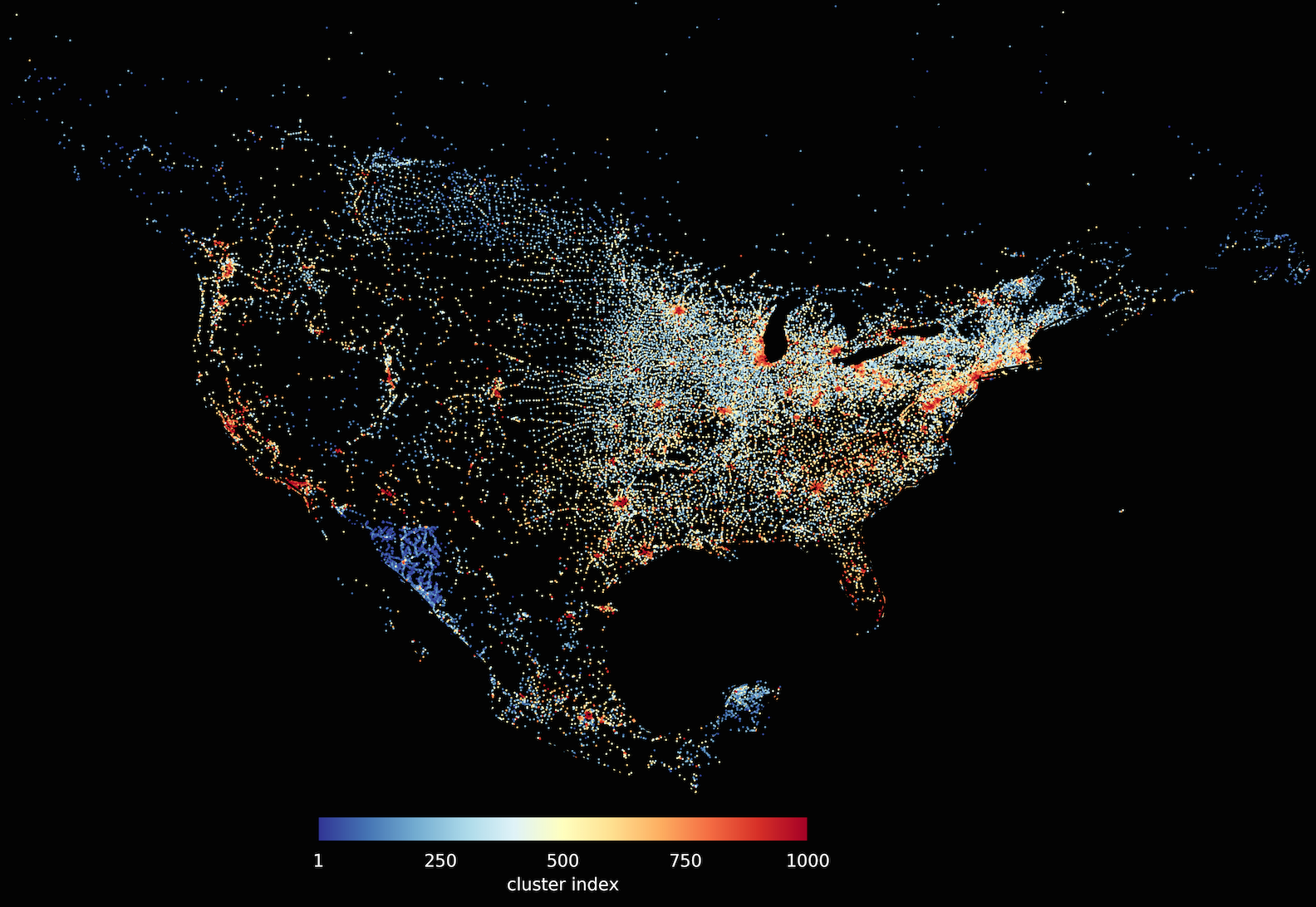} \caption{Distribution of 33,875 urban forms in North America: Similar colors refer to similar urban forms that all together they create meaningful spatial patterns of development around main metropolitans} 
\label{fig:Fig10} \end{figure}

We trained a one-dimensional SOM with 2000 nodes and all the urban vectors as the training data set. The first cluster always automatically collects those locations that do not have a lot of spatial information or are very sparse. Therefore, we removed the first cluster from all our analyses. Figure \ref{fig:Fig7} shows the distribution of urban forms within the identified clusters. As expected, this distribution has a long tail, which corresponds to unique cities with high levels of density. Further there are two peaks in the distribution, which are reflecting the most common types of urban forms, corresponding to towns and villages mainly. 
Using the trained one-dimensional SOM we can easily render the urban forms in a single geo-map, where similar urban forms get similar colors based on their assigned cluster numbers\cite{moosavi2017contextual}. In this way, we can also visualize the spatial (dis-) similarity of urban forms. As Figure \ref{fig:Fig8} shows the first peak in the distribution of the urban forms (starting from blue color) corresponds to villages or underdeveloped areas. Further, the majority of towns and small cities in developed areas are placed around the second peak of the distribution. Further, unique cities and very dense cities like the ones in Japan are placed on the right tale of the distribution of urban forms and get a red color on the map. 
In usual visualizations of low dimensional grids of SOM, (e.g. in Figure \ref{fig:Fig6}), all the clusters are uniformly rendered in a rectangular grids. However, inspired by Topological Data Analysis (TDA)\cite{carlsson2009topology}, one can study the topological shape of urban patterns by considering the degree of similarity between each pairs of clusters. By setting a threshold in the similarity between two nodes of SOM we can represent the trained SOM as a graph, where if the similarity of two nodes are more than the threshold, there is an edge between two nodes. Similar to the idea of persistence diagram in TDA, by changing the similarity thresholds, we can identify the most persistent shape of this graph. We tested several similarity thresholds with different cluster sizes. Finally, with similarity threshold of 80 percent (based on normalized Euclidean similarities) and 2000 clusters, Figure \ref{fig:Fig9} shows the topology of urban forms, considering all the locations over the planet. In this diagram each node corresponds to one cluster of the trained SOM. The size of each node is proportional to the number of unique urban forms within that cluster. As it is shown, there is a ring naturally formed by the clusters on one side of the spectrum, followed by a very long tale to the other extreme of urban forms, which are expanded to different unique clusters with few members. The colors of the nodes correspond to the cluster labels in the same way that is shown in Figure \ref{fig:Fig8}. 

Further, it is worth to mention that the learned urban vectors can be used for comparison of any selection of urban forms or in general can be transferred for any other analysis. For instance, using the learned vectors of urban forms limited to North America, we trained another SOM to visualize the spatial patterns of development in this region. Figure \ref{fig:Fig10} shows the distribution of 33,875 cities, towns and villages in North America, where similar locations with similar color codes creates meaningful spatial patterns of development around main metropolitans. 

\section*{Discussion} Availability of large spatial data sets across the planet such as OSM offers new opportunities for the study of urban development patterns all over the world. However, most of the traditional approaches do not cope with these large data sets. Majority of existing studies on urban form and urban morphology are based on limited observations and qualitative approaches that are not easily reproducible or lack consistent and easy to use research tools. 

As an alternative to the traditional methods, in this work we used a deep neural network that automatically learns and encodes the complex and hierarchical patterns of urban forms all over the world. We trained a deep convolutional auto encoder over the images of street networks in around one million cities, towns and villages. The trained model encodes the images of the urban forms to vector based representations, while it considers the overall spatial structure and other factors such as orientation, graphical structure, density and partial deformations of urban forms. The learned urban vectors were used for searching, comparing and finding similar urban patterns. Further, we developed an interactive spectrum of urban forms that can be easily explored by urban planners and researchers. Further, by analysis of clusters of urban forms and using the ideas from topological data analysis we investigated the distribution of the current urban forms all over the planet.

These results are among very early outputs of an ongoing project, where by coupling large multimodal urban data sets with machine learning algorithms, new ways of comparative research at the global scale is becoming feasible. The long-term goal of this project is to develop a \emph{city explorer} and a \emph{search engine of cities}, where depending on a specific goal (e.g. a certain question about a specific city), one can find similar developments somewhere in the world. Thinking about this approach in the same way that people use Internet as a complex and dynamic library, it can introduce new ways of working with complex urban phenomena. In this regard, the role of machine learning is very essential as a powerful “mediator” that produces very useful extracts from multimodal Big Data collections, which is definitely not possible to address by classical approaches.

In addition to explorative analyses of urban forms, it is also possible to ask targeted questions related to urban forms and functions. For instance, one can simultaneously consider the effect of urban form on urban indicators in terms of economy, health, environment, and transportation quality in cities. As an example, American Community Survey (ACS) with a lot of information on more than 60K locations (census tracts) in USA is an immediate Big Data for training several multimodal deep-learning models. In the next step we will investigate the performance of these types of supervised models.

\section*{Data availability} All the collected data and the codes to generate the results are available from \href{https://sevamoo.github.io/cityastext}{the project website}.

\bibliography{sample}

\end{document}